\documentclass[10pt,twocolumn,letterpaper]{article}

\usepackage[pagenumbers]{cvpr} 

\usepackage[dvipsnames]{xcolor}

\definecolor{cvprblue}{rgb}{0.21,0.49,0.74}
\usepackage[pagebackref,breaklinks,colorlinks,citecolor=cvprblue]{hyperref}

\usepackage[utf8]{inputenc} 
\usepackage[T1]{fontenc}    

\usepackage{graphicx}
\usepackage{amsmath}
\usepackage{amssymb}
\usepackage{bbm}

\usepackage{cite}
\usepackage{amsmath,amssymb,amsfonts}
\usepackage{algorithmic}
\usepackage{textcomp}
\usepackage{subcaption}
\usepackage{tabularx}
\usepackage{booktabs}       

\usepackage{amsmath}
\usepackage{amssymb}
\usepackage{bm}
\usepackage{color}
\usepackage{xcolor,soul}
\usepackage{algorithm}
\usepackage{multirow}
\usepackage{xspace}
\usepackage{soul}

\usepackage{enumitem}

\usepackage{newtxtext}
\newdimen\abovecrulesep
\newdimen\belowcrulesep
\makeatletter
\patchcmd{\@@@cmidrule}{\aboverulesep}{\abovecrulesep}{}{}
\patchcmd{\@xcmidrule}{\belowrulesep}{\belowcrulesep}{}{}

\definecolor{demphcolor}{RGB}{144, 144, 144}
\definecolor{mygray}{gray}{0.4}
\definecolor{lightgray}{rgb}{0.9, 0.9, 0.9}
\definecolor{deepgreen}{RGB}{0,100,0}
\newcommand{\demph}[1]{\textcolor{demphcolor}{#1}}

\newlength\savewidth
\newcommand\shline{\noalign{\global\savewidth\arrayrulewidth\global\arrayrulewidth 1pt}\hline\noalign{\global\arrayrulewidth\savewidth}}

\newcommand{\tablestyle}[2]{\setlength{\tabcolsep}{#1}\renewcommand{\arraystretch}{#2}\centering\small}
\makeatletter\renewcommand\paragraph{\@startsection{paragraph}{4}{\z@}{.5em\@plus1ex\@minus.2ex}{-.5em}{\normalfont\normalsize\bfseries}}
\makeatother

\newcommand{\modelname}{mPLUG-Owl2\xspace}

\setstcolor{blue}

\usepackage[capitalize]{cleveref}
\crefname{section}{Sec.}{Secs.}
\Crefname{section}{Section}{Sections}
\Crefname{table}{Table}{Tables}
\crefname{table}{Tab.}{Tabs.}

\def\confName{CVPR}
\def\confYear{2024}

\title{mPLUG-Owl2: Revolutionizing Multi-modal Large Language Model \\ with Modality Collaboration}

\author{
Qinghao Ye\thanks{Equal contribution}\hspace{6mm} Haiyang Xu\footnotemark[1]\hspace{6mm} Jiabo Ye\footnotemark[1]\hspace{6mm} Ming Yan\thanks{Corresponding author}\hspace{6mm} Anwen Hu \\ Haowei Liu\hspace{6mm} Qi Qian\hspace{6mm} Ji Zhang\hspace{6mm} Fei Huang\hspace{6mm} Jingren Zhou\\
Alibaba Group \\
{\tt\small \{yeqinghao.yqh, shuofeng.xhy, yejiabo.yjb, ym119608\}@alibaba-inc.com} \\
{\small Code \& Demo \& Models: \ \ \url{https://github.com/X-PLUG/mPLUG-Owl/tree/main/mPLUG-Owl2}}
}
\begin{document}
\maketitle
\begin{abstract}
Multi-modal Large Language Models (MLLMs) have demonstrated impressive instruction abilities across various open-ended tasks. However, previous methods primarily focus on enhancing multi-modal capabilities. In this work, we introduce a versatile multi-modal large language model, \modelname, which effectively leverages modality collaboration to improve performance in both text and multi-modal tasks. \modelname utilizes a modularized network design, with the language decoder acting as a universal interface for managing different modalities. Specifically, \modelname incorporates shared functional modules to facilitate modality collaboration and introduces a modality-adaptive module that preserves modality-specific features. Extensive experiments reveal that \modelname is capable of generalizing both text tasks and multi-modal tasks and achieving state-of-the-art performances with a single generic model. Notably, \modelname is the first MLLM model that demonstrates the modality collaboration phenomenon in both pure-text and multi-modal scenarios, setting a pioneering path in the development of future multi-modal foundation models.
\end{abstract}

\section{Introduction}

\begin{figure}[h]
    \centering
     \vspace{-2ex}
    \includegraphics[width=1\linewidth]{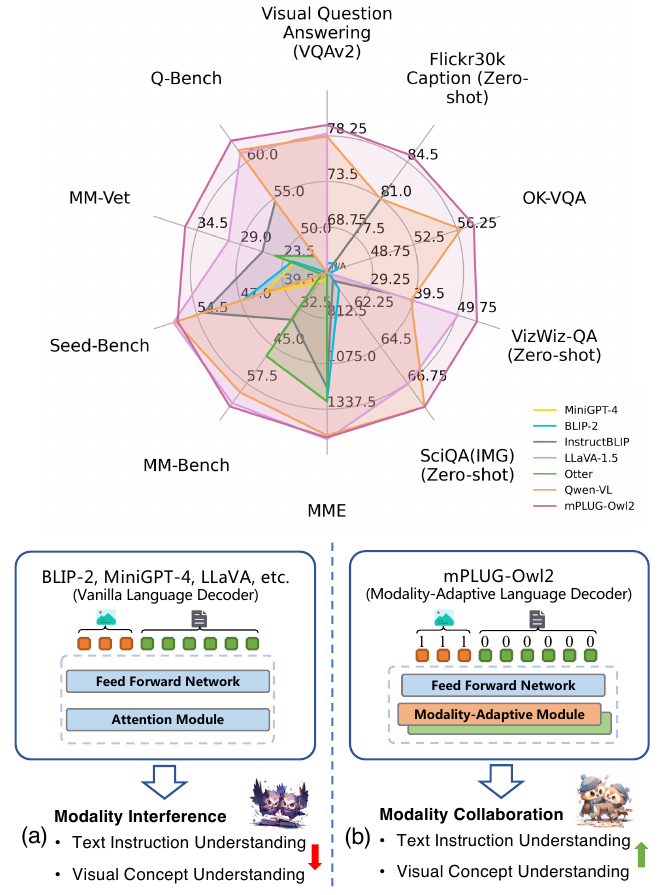}
    \caption{An overall performance comparison between \modelname and existing MLLMs and difference between existing MLLMs and our proposed model. (a) Previous approaches utilize a standard language decoder (i.e., LLM) to manage different types of instructions, leading to modality interference and performance degradation. (b) We introduce \modelname, which uses a modality-adaptive language decoder to handle different modalities within distinct modules while sharing some parameters for modality collaboration. This approach mitigates the issue of modality interference.}
    \vspace{-2ex}
    \label{fig:modality-colab}
\end{figure}

Large Language Models (LLMs) such as GPT-3 \cite{Brown2020gpt3}, LLaMA \cite{Touvron2023LLaMA, Touvron2023Llama2}, and GPT-4 \cite{OpenAI2023gpt4} have garnered significant attention due to their exceptional generalization abilities in text understanding and generation. To facilitate the vision-language applications, GPT-4V\footnote{https://openai.com/research/gpt-4v-system-card} \cite{2023GPT4VisionSC} has recently demonstrated impressive multi-modal capabilities in diverse tasks, e.g., description , question answering, etc., sparking interest among researchers in the potential convergence of the vision-language field. This has led to the emergence of a group of Multi-modal Large Language Models (MLLMs) \cite{Zhu2023MiniGPT4, Liu2023Llava, ye2023mplugowl, mplugdocowl, ye2023ureader, Bai2023QwenVL, Dai2023InstructBLIP, Li2023BLIP2}, which aim to enhance LLMs with the ability to understand and handle visual problems.

Previous studies \cite{xu2023mplug2, kwon2022masked} in multi-modal learning suggest that different modalities can effectively collaborate, thereby enhancing the performance of both text and multi-modal tasks simultaneously. However, MLLMs is a unified model that supports different modalities and tasks without fine-tuning for specific tasks. Recent works utilize cross-modal alignment modules (e.g., Q-former \cite{Zhu2023MiniGPT4, Dai2023InstructBLIP, Li2023BLIP2} and linear layer \cite{Liu2023Llava, Chen2023Shikra}) to map visual features from the vision encoder into the frozen LLMs to carry out multi-modal tasks by leveraging preserved language capabilities. This strategy, unfortunately, restricts the potential of modality collaboration. As a result, some researchers \cite{Liu2023Llava, ye2023mplugowl} opt to fine-tune LLMs during multi-modal instruction tuning. While fine-tuning significantly improves multi-modal tasks, it risks weakening text task performance \cite{Driess2023PaLME}. As illustrated in Figure \ref{fig:modality-colab}, the challenge of modality collaboration in MLLMs is from applying a single module to balance the gain of modality collaboration and modality interference, where modalities may interfere with each other on a large number of instruction datasets across multiple modalities.

To mitigate this challenge, we present a new general-purpose multi-modal foundation model, \modelname, in this work. Our model features a modularized network design that takes both modality collaboration and modality interference into account, using the language decoder as a universal interface for managing multi-modal signals. Specifically, \modelname incorporates certain shared functional modules to promote modality collaboration and introduces a modality-adaptive module that serves as a pivot across different modalities. Therefore, vision and language modalities are projected into a shared semantic space for cross-modality interaction, while the proposed module helps preserve modality-specific features. With our novel architecture, modalities with varying information densities are shielded from modality interference due to the modality-adaptive module and can collaborate effectively in capturing shared information. Furthermore, we introduce an innovative two-stage training paradigm that consists of vision-language pre-training and joint vision-language instruction tuning. This paradigm trains the vision encoder across two stages, enabling it to capture both low-level and high-level semantic visual information more effectively.

Extensive experiments illustrate the effectiveness and generalization abilities of \modelname, which achieves state-of-the-art performance on 8 classic vision-language benchmarks using \textbf{a single generic model}. Furthermore, it either first or second in performance on 5 recent zero-shot multi-modal benchmarks, underscoring its adaptability and proficiency in multi-modal instruction comprehension and generation. In addition to its cutting-edge performance in multi-modal tasks, \modelname also achieves state-of-the-art results on multiple pure-text benchmarks. Moreover, we provide in-depth analysis to demonstrate and validate the impact of modality collaboration through our proposed modality-adaptive module, especially in enhancing text tasks, including understanding, knowledge, and reasoning. Finally, comprehensive ablation studies validate the effectiveness of the proposed MLLM training paradigm, which can help inspire the development of future multi-modal foundation models.

\vspace{-2ex}
\section{Related Work}
\vspace{-1ex}
\paragraph{Multi-Modal Large Language Foundation Models.}
The successful application of Large Language Models (LLMs) has paved the way for developing several approaches aiming to augment the perceptual capacities of LLMs with additional modalities, all within a unified model. There are three primary methods for constructing multi-modal large language foundational models, each showing promise for robust zero-shot generalization capabilities in the vision-language domain. For instance, Flamingo \cite{alayrac2022flamingo} is a forerunner in this area, using a frozen vision encoder and a large language model equipped with gated cross-attention for cross-modality alignment. In contrast, PaLM-E \cite{Driess2023PaLME} integrates extracted visual features directly through linear layers into the pre-trained PaLM \cite{Chowdhery2022PaLM} model, which boasts 520 billion parameters, thereby leading to robust performance across numerous real-world applications. This approach has been broadly adopted by models such as LLaVA \cite{Liu2023Llava}, Shikra \cite{Chen2023Shikra}, etc. One significant limitation of this method, however, is the creation of lengthy visual sequences. To address this, BLIP-2 \cite{Li2023BLIP2}, drawing inspiration from DETR \cite{carion2020detr}, developed a Q-former to reduce the sequence length of visual features efficiently. This design has been mirrored by Kosmos-1 \cite{Huang2023Kosmos1}, mPLUG-Owl \cite{ye2023mplugowl}, and MiniGPT-4 \cite{Zhu2023MiniGPT4}. Nevertheless, it should be noted that these methods directly align the visual features with the LLMs, treating vision and language signals as equivalent, thereby overlooking the unique granularities between vision and language modalities. To alleviate this problem, we introduce modality-adaptive module. Our proposed model leads to superior performance in both zero-shot and fine-tuning evaluation settings in terms of both image and video.

\vspace{-1ex}
\paragraph{Instruction Tuning with MLLMs.}
Instruction tuning optimizes pre-trained large language models to comprehend and adhere to natural instructions, thereby enhancing their ability to generalize unseen tasks in a zero-shot manner. Researchers often employ models such as ChatGPT and GPT-4 \cite{OpenAI2023gpt4} to generate diverse and expansive instruction datasets, including those like Alpaca \cite{alpaca}, ShareGPT \cite{ShareGPT2023}, and WizardLM \cite{Xu2023WizardLM}. As multi-modal large language models emerge, research communities are beginning to create high-quality, diverse multi-modal datasets. For instance, MiniGPT-4 \cite{Zhu2023MiniGPT4} utilizes GPT-3.5 to rephrase captions generated by pre-trained models. Concurrently, LLaVA \cite{Liu2023Llava}, SVIT \cite{zhao2023svit}, and LRV-Instruction \cite{liu2023lrv} take advantage of image annotations, such as bounding boxes of objects, image captions, and region descriptions, to prompt GPT-4 to generate instructions and responses using self-instruction methods. Models such as mPLUG-Owl \cite{ye2023mplugowl} and LLaVA-1.5 \cite{Liu2023LLava15} further advance this area by undergoing joint training with language-only and vision-and-language instruction data, thereby mitigating the risk of catastrophic forgetting of language knowledge. Rather than merely preventing this phenomenon of catastrophic forgetting, \modelname, with the help of the modality-adaptive module, can gain from the collaborative efforts of modalities by being jointly trained with language-only and multi-modal instruction data, thus enhancing both multi-modal and language-only performance.

\begin{figure*}[!htbp]
    \centering
    \includegraphics[width=1\textwidth]{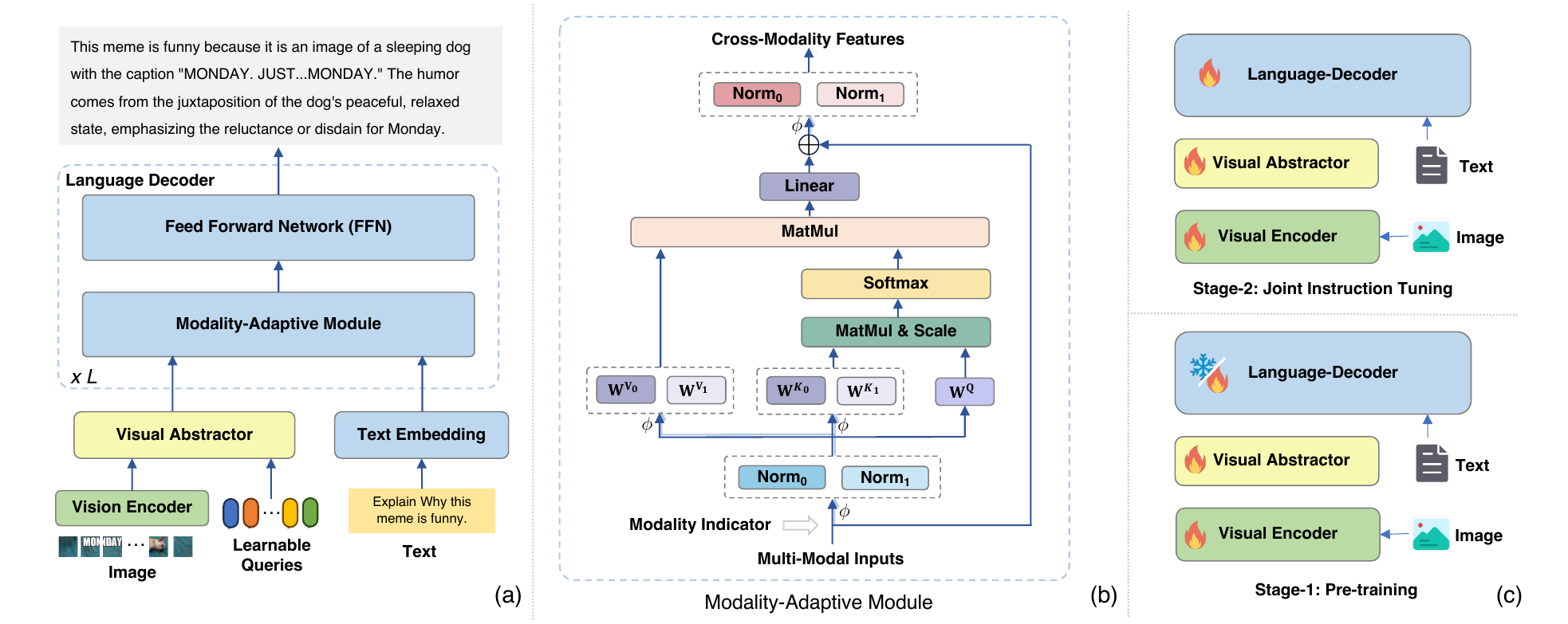}
    \vspace{-1ex}
    \caption{Illustration of the proposed \modelname and its training paradigm. (a) An overview of \modelname, which consists of a vision encoder, visual abstractor, text embedding layer, and a language decoder. (b) Details of the proposed modality-adaptive module, which takes multi-modal inputs and employs different parameters to project various modalities into a shared semantic space for relational learning while preserving modality-specific features, thereby enabling modality collaboration. (c) The training paradigm of \modelname involves first pre-training the visual-related modules, including the vision encoder and visual abstractor. Simultaneously, newly added parameters in the language decoder are also learned during the pre-training stage. During the instruction tuning stage, both language instructions and multi-modal instructions are used to jointly train the entire model.}
    \vspace{-2ex}
    \label{fig:arch}
\end{figure*}

\vspace{-1ex}
\section{Methodology}
\subsection{Overview}
Figure \ref{fig:arch} (a) sketches the overview of the \modelname. Specifically, our model comprises a vision encoder, a visual abstractor, a text embedding layer, and a language decoder. Notably, the standard implementation of the text embedding layer and language decoder involves the use of a large language model, such as GPT \cite{Brown2020gpt3} or LLaMA \cite{Touvron2023LLaMA}. We first briefly introduce our model's architecture in Section \ref{sec:arch}. Furthermore, we handle different types of modalities by introducing the modality-adaptive module in Section \ref{sec:module}. Lastly, we introduce the training paradigm for training \modelname with modality collaboration in Section \ref{sec:training}.

\subsection{Model Architecture}\label{sec:arch}
As depicted in Figure \ref{fig:arch}, our model, referred to as \modelname, is composed of three main components: a fundamental vision encoder \cite{radford2021clip}, a visual abstractor, and a language decoder. Specifically, we utilize ViT-L/14 as the vision encoder and LLaMA-2-7B \cite{Touvron2023Llama2} as the language decoder. The vision encoder processes an input image with an $H\times W$ resolution and produces a sequence of $\frac{H}{14} \times \frac{W}{14}$ tokens. These visual token features are then combined with text token embeddings and fed into the language decoder that serves as a universal interface that converts various vision-language tasks into text-generation tasks. 
However, with the increase in image resolution, the encoded visual token sequences can exponentially lengthen. Additionally, the presence of abundant redundancy in the images (e.g., background, similar patches) leads to computational waste and introduces considerable noise. To address this, we propose a visual abstractor equipped with a fixed set of learnable queries to extract higher semantic features from images. Specifically, we feed the extracted visual token sequence $\mathcal{I} = [I_1, I_2, \cdots, I_P] \in \mathbb{R}^{P\times d}$ and a fixed number of $K$ learnable queries $\mathcal{Q} \in \mathbb{R}^{K\times d}$ into the visual abstractor. Here, $P=\frac{H}{14} \times \frac{W}{14}$ represents the number of visual patches, and $D$ is the hidden dimension. The visual abstractor consists of a series of visual abstractor layers. In the $i$-th layer of the visual abstractor, the compressed visual representations $\mathcal{V}^{i+1}$ are computed as follows:
\begin{align}
    \mathcal{C}^{i} &= Attn(\mathcal{V}^{i}, [\mathcal{I}; \mathcal{V}^{i}], [\mathcal{I}; \mathcal{V}^{i}]), \\
    \mathcal{V}^{i+1} &= SwiGLU(\mathcal{C}^{i}W_1)W_2.
\end{align}
Here, $Attn(\cdot, \cdot, \cdot)$ represents the self-attention operation, while $W_1 \in \mathbb{R}^{d\times d'}$ and $W_2 \in \mathbb{R}^{d'\times d}$ are learnable parameters. The function $SwiGLU(\cdots)$ refers to the SwiGLU activation function \cite{shazeer2020swiglu}. We designate $\mathcal{V}^{0} = \mathcal{Q}$ to initiate the process. Moreover, to augment the fine-grained perception ability, we integrate sinusoidal positional embeddings with the image feature $\mathcal{I}$ and $\mathcal{V}^{i}$, thereby preserving positional information, which has been proven essential in \cite{carion2020detr}. Hence, the computation required by the language decoder decreases from $O((P+L)^2)$ to $O((K+L)^2)$, significantly reducing computational load when $P \gg K$, particularly in scenarios involving multiple images and when the text length $L$ is relatively short. Once the compressed visual feature is obtained, it is concatenated with text token embeddings and then processed by the language decoder to generate the prediction.

\subsection{Modality-Adaptive Module}\label{sec:module}
Prior approaches \cite{Zhu2023MiniGPT4, Liu2023Llava, Dai2023InstructBLIP, ye2023mplugowl} typically attempt to align visual features with language features by projecting image features into the language semantic space. However, this strategy can cause a mismatch in granularity , where image features often contain fruitful semantic information compared to the discrete semantic information within text embedding features. Those methods disregard the unique characteristics of visual and textual information, thus potentially limiting the model's performance. To this end, we propose a new approach, namely, the Modality-Adaptive Module (MAM), which decouples vision-language representations by projecting visual features and language features into a shared semantic space while preserving the distinctive properties of each modality. 

Formally, given a vision-language sequence $X \in \mathbb{R}^{(L_V + L_T) \times d}$ and modality indicators $M \in {\{0, 1\}}^{(L_v + L_T)}$, we first define modality separated operation $\phi$ as:
\begin{align}
    \phi(X, M, m) = X \odot \mathbbm{1}_{\{M = m\}},
\end{align}
where $m \in \{0,1\}$ is the type of modalities (i.e., vision or language). Given the previous layer's output vectors $H_{l-1}, l\in[1, L]$, where $L$ is the number of language decoder layers, we first normalized different modalities into the same magnitude as follows:
\begin{equation}
    \small
    \tilde{H}_{l-1} = LN_V(\phi(H_{l-1}, M, 0)) + LN_T(\phi(H_{l-1}, M, 1)),
\end{equation}
where $LN_V$ and $LN_T$ are layer normalization \cite{ba2016layer} for visual features and language features respectively. Then, we reformulate the self-attention operation by leveraging separated linear projection layers for key projection matrix and value projection matrix while preserving query projection matrix shared as follows:
\begin{align}
    H^Q_{l} &= \tilde{H}_{l-1} W^Q_l, \\ 
    H^K_{l} &= \phi(\tilde{H}_{l-1}, M, 0) W^{K_0}_l + \phi(\tilde{H}_{l-1}, M, 1) W^{K_1}_l, \\
    H^V_{l} &= \phi(\tilde{H}_{l-1}, M, 0) W^{V_0}_l + \phi(\tilde{H}_{l-1}, M, 1) W^{V_1}_l, \\
    C_{l} &= Softmax\left(\frac{H^Q_{l} {H^K_{l}}^\top}{\sqrt{d}}\right)H^V_{l},
\end{align}
where $W^Q_l, W^{K_0}_l, W^{K_1}_l, W^{V_0}_l, W^{V_1}_l \in \mathbb{R}^{d\times d}$ are the learnable projection matrices, and $C_l \in \mathbb{R}^{(L_V + L_T) \times d}$ is the context features of $l$-th layer. In this manner, we can calculate the similarities between these two modalities within a shared semantic space, while also preserving the unique characteristics of each modality through different value projection layers. Moreover, by decoupling the key and value projection matrix, we can avoid interference between the two modalities, particularly in relation to granularity mismatch. In a similar vein, we also aim to model these characteristics by using different layer normalization layers. Finally, in order to promote modality collaboration within the same feature space, we maintain a shared FFN for both modalities. 
As a consequence, the model is able to preserve modality characteristics while achieving modality collaboration via the proposed modality-adaptive module.

\begin{table*}
\centering
    \tablestyle{4pt}{1.1} 
    \def \w{20pt} 
    \resizebox{0.85\linewidth}{!}{
    \begin{tabular}{c|l|c|cc|ccc|ccc}
        \shline
        ~ & ~ & ~ & \multicolumn{2}{c}{Image Caption} & \multicolumn{3}{c}{General VQA} & \multicolumn{3}{c}{General VQA (Zero-shot)}  \\
        \cmidrule(lr){4-5} \cmidrule(lr){6-8} \cmidrule(lr){9-11}
        \multirow{2}{*}{Model Type} & \multirow{2}{*}{Method} & \multirow{2}{*}{\#Params} & \multirow{2}{*}{COCO} & Flickr30K & \multirow{2}{*}{VQAv2} & \multirow{2}{*}{OKVQA} & \multirow{2}{*}{GQA} & \multirow{2}{*}{VizWizQA} & \multirow{2}{*}{TextVQA} & \multirow{2}{*}{SciQA (IMG)} \\
        ~ & ~ & ~ & ~ &  (Zero-Shot) & ~ & ~ & ~ & ~ & ~ & ~ \\
        \hline
        \multirow{9}{*}{Generalists} & BLIP-2 \cite{Li2023BLIP2} & 8.2B & - & 74.9 & 65.0 & 45.9 & 41.0 & 19.6 & 42.5 & 61.0 \\
        & InstructBLIP \cite{Dai2023InstructBLIP} & 8.2B & 102.2 & 82.4 & - & - & 49.2 & 34.5 & $50.1^\dag$ & 60.5 \\
        & Unified-IO$_{\text{XL}}$ \cite{Lu2022UnifiedIO} & 2.9B & 122.3 & - & 77.9 & 54.0 & - & \demph{$57.4^\ddag$} & - & - \\
        & PaLM-E-12B \cite{Driess2023PaLME} & 12B & 135.0 & - & 76.2 & 55.5 & - & - & - & - \\
        & Shikra \cite{Chen2023Shikra} & 7.2B & 117.5 & 73.9 & 77.4 & 47.2 & - & - & - & - \\
        & LLaVA-1.5 \cite{Liu2023LLava15} & 7.2B & - & - & 78.5 & - & \textbf{62.0} & 50.0 & 46.1/$58.2^\dag$ & 66.8 \\
        & Qwen-VL-Chat \cite{Bai2023QwenVL} & 9.6B & 131.9 & 81.0 & 78.2 & 56.6 & 57.5 & 38.9 & \demph{$61.5^\ddag$} & 68.2 \\
        \cmidrule(lr){2-11}
        & \textbf{\modelname} & 8.2B & \textbf{137.3} & \textbf{85.1} & \textbf{79.4} & \textbf{57.7} & 56.1 & \textbf{54.5} & \textbf{54.3/$\textbf{58.2}^\dag$} & \textbf{68.7} \\
        \hline
        \multirow{3}{*}{\demph{Specialists}} & \demph{GIT} \cite{wang2022git} & 0.7B & \demph{114.8} & \demph{49.6} & \demph{78.6} & \demph{-} & \demph{-} & \demph{68.0} & \demph{59.8} & \demph{-} \\
        & \demph{GIT2} \cite{wang2022git} & 5.1B & \demph{145.0} & \demph{50.7} & \demph{81.7} & \demph{-} & \demph{-} & \demph{71.0} & \demph{59.8} & \demph{-} \\
        & \demph{PaLI-17B} \cite{chen2022pali} & 17B & \demph{149.1} & \demph{-} & \demph{84.3} & \demph{64.5} & \demph{-} & \demph{71.6} & \demph{58.8} & \demph{-} \\
        \shline
    \end{tabular}
    }
    \caption{\textbf{Performance comparison on image caption and visual question answering.} For image caption, CIDEr is reported for evaluation, and accuracy is reported for VQA. Note that specialists are fine-tuned on each individual dataset. \dag\ denotes OCR inputs are utilized. \ddag\ indicates the model has trained on the dataset. We gray out those specialists' methods which are individually fine-tuned on the dataset as well as those fine-tuned results of generalists.
    }
    \label{table:multimodal-results}
\end{table*}

\begin{table*}
\centering
    \tablestyle{7pt}{1.1} 
    \def \w{15pt}
    \resizebox{0.9\linewidth}{!}{
    \begin{tabular}{l|c|c|c|c|c|c|c}
        \shline
        Method    & Vision Encoder & Language Model & MME     & MMBench & MM-Vet & SEED-Bench & Q-Bench \\
        \hline
        BLIP-2 \cite{Li2023BLIP2}      & ViT-g (1.3B)            & Vicuna (7B)             & 1293.84 & -       & 22.4   & 46.4       & -       \\
        MiniGPT-4 \cite{Zhu2023MiniGPT4}  & ViT-g (1.3B)            & Vicuna (7B)             & 581.67  & 23.0    & 22.1   & 42.8       & -       \\
        LLaVA \cite{Liu2023Llava}      & ViT-L (0.3B)            & Vicuna (7B)             & 502.82  & 36.2    & 28.1   & 33.5       & 54.7   \\
        mPLUG-Owl \cite{ye2023mplugowl} & ViT-L (0.3B)    & LLaMA (7B) & 967.34 & 46.6 & - & 34.0 & 58.9 \\
        InstructBLIP \cite{Dai2023InstructBLIP}  & ViT-g (1.3B)            & Vicuna (7B)             & 1212.82 & 36.0    & 26.2   & 53.4       & 55.8   \\
        LLaMA-Adapter-v2 \cite{Gao2023LLaMAAdapterV2} & ViT-L (0.3B)            & LLaMA (7B)              & 1328.40 & 39.5    & 31.4   & 32.7       & 58.1   \\
        Otter \cite{Li2023Otter}           & ViT-L (0.3B)            & LLaMA (7B)              & 1292.26 & 48.3    & 24.6   & 32.9       & 47.2   \\
        Qwen-VL-Chat \cite{Bai2023QwenVL}    & ViT-G (1.9B)            & Qwen (7B)               & 1487.58 & 60.6    & -      & 58.2       & 61.6   \\
        LLaVA-1.5 \cite{Liu2023LLava15}     & ViT-L (0.3B)            & Vicuna (7B)             & \textbf{1510.70} & 64.3    & 30.5   &  \textbf{58.6}       & 60.7   \\        
        \hline
        \textbf{\modelname} & ViT-L (0.3B)   & LLaMA (7B) & 1450.19 & \textbf{64.5} & \textbf{36.2} & 57.8 & \textbf{62.9} \\
        \shline
    \end{tabular}
    }
    \caption{\textbf{Zero-shot multi-modal evaluation on multi-modal benchmarks} including MME \cite{fu2023mme}, MMBench \cite{liu2023mmbench}, MM-Vet \cite{yu2023mmvet}, SEED-Bench \cite{li2023seedbench}, and Q-Bench \cite{wu2023qbench}. The overall scores are reported for evaluation. For MMBench and Q-Bench, we report test results.}
    \vspace{-2ex}
    \label{table:zeroshot-multimodal-bench}
\end{table*}

\subsection{Training Paradigm}\label{sec:training}
As depicted in Figure \ref{fig:arch} (c), we employ a two-stage approach in training \modelname, comprising pre-training and visual instruction tuning similar to \cite{Liu2023Llava, ye2023mplugowl}, which aims to align the pre-trained vision encoder and language model during the pre-training phase, and then fine-tune the language model with language modeling loss during the instruction tuning phase. However, we find that simply freezing a pre-trained vision encoder and training a vision-language projector to align visual data with language models can limit their capacity to interpret complex visual information, such as scene text and visual knowledge. To address the issue, we make the vision encoder trainable throughout both the pre-training and instruction tuning stages. This strategy allows the model to capture both low-level and high-level semantic visual information more effectively. Specifically, for the pre-training stage, we enable the vision encoder, visual abstractor, and a part of the modality-adaptive module to be trainable, while keeping the pre-trained language model frozen. 
Meanwhile, prior research in multi-modal learning \cite{xu2023mplug2} has indicated that significant enhancements can be achieved through the collaborative learning of uni-modal and multi-modal sources. Based on this, we adopt a joint training approach by tuning the whole model during the instruction tuning stage, incorporating both text and multi-modal instructions. This methodology enhances the model's comprehension of visual concepts embedded within the text by the multi-modal instructions. Concurrently, the text instruction data augments the model's understanding of intricate natural instructions, thereby ensuring the preservation of its linguistic capabilities.

\section{Experiments}
\subsection{Implementation}
\paragraph{Data sets} 
\modelname is first pre-trained on image-text pairs and fine-tunes on mono-modal and multi-modal instruction data. For pre-training data, we randomly pick about 400 million image-text pairs from five public datasets: Conceptual Captions (CC3M/CC12M) \cite{changpinyo2021cc3m12m}, COCO \cite{lin2014coco}, Laion-en \cite{schuhmann2022laion}, COYO \cite{kakaobrain2022coyo-700m}, DataComp \cite{gadre2023datacomp}. For instruction data, we collect 5 types of datasets including 1) image captioning (i.e., TextCaps \cite{sidorov2020textcaps}, COCO \cite{lin2014coco}); 2) image question answering (i.e., VQAv2 \cite{balanced_vqa_v2}, OKVQA \cite{marino2019okvqa}, OCR-VQA \cite{mishra2019ocrvqa}, GQA \cite{hudson2019gqa}, and A-OKVQA \cite{schwenk2022aokvqa}); 3) region-aware QA (i.e., RefCOCO \cite{yu2016refcoco}, VisualGenome \cite{krishna2017visualgenome}); 4) multi-modal instruct data (i.e., LLaVA-instruct-150K \cite{Liu2023Llava}); 5) text-only instruct data (i.e., ShareGPT-80K \cite{ShareGPT2023}, SlimOrca \cite{SlimOrca}). Details can be found in the Appendix.

\paragraph{Training Settings}
We pre-train the model for 42,500 iterations with a batch size 8,192 for about 348 million image-text pairs. Since we adopt the language modeling loss, the large batch size can be easily achieved by the gradient accumulation technique. \modelname adopts ViT-L \cite{radford2021clip} with patch size $14\times 14$ and pre-trained at resolution $224 \times 224$. We use the same data augmentation in BLIP-2 \cite{Li2023BLIP2}, including random resized cropping, and horizontal flipping with a probability of 0.5. The number of layers in the visual abstractor is set to 6 and it is randomly initialized. The number of learnable queries is set to 64. For the language model, LLaMA-2 \cite{Touvron2023Llama2} is employed for handling multi-modal features with 7B parameters, and the parameters of modality-adaptive modules are initialized from the language model. We use the AdamW \cite{loshchilov2018adamW} optimizer with $\beta_1 = 0.9$, $\beta_2 = 0.98$ and $\epsilon=$1e-6 for optimization. The cosine learning rate decay scheduler with a peak learning rate of 1e-4 and with warmup steps 1k. For the learning rate of the vision encoder, we employ layer-wise learning rate decay with a factor of 0.9 to retain the low-level visual representation. For the instruction tuning stage, we train the whole model for 1 epoch with a learning rate of 2e-5 and batch size 256. Besides, we increase the resolution from $224 \times 224$ to $448 \times 448$. The layer-wise learning rate decay is also employed which is crucial for retaining good visual representation in our experiments.

\subsection{Main Results}
\paragraph{Image Caption and Visual Question Answering.}
We assess \modelname using a wide range of academic benchmarks for evaluating vision-language models. Our evaluation includes eight popular benchmarks, as summarized in Table \ref{table:multimodal-results}. As the results show, our \modelname surpasses previous generalist models in both captioning and question answering tasks. Specifically, \modelname achieves state-of-the-art performance on the Flickr30K datasets, even compared with models with more powerful backbones (e.g., Qwen-VL-Chat \cite{Bai2023QwenVL} and InstructBLIP \cite{Dai2023InstructBLIP}). Moreover, \modelname exhibits distinct advantages in visual question answering, especially in OCR-free scenarios, where \modelname achieves 54.3\% accuracy on the TextVQA dataset in a zero-shot manner, demonstrating the benefits of our training strategy. Also worth noting is that \modelname shows strong zero-shot performance on the ScienceQA (Image Set) and VizWizQA datasets.

\paragraph{MLLM-oriented Multi-modal Benchmarks.} 
Given the robust zero-shot capabilities of Multi-Modal Language Models (MLLMs), traditional evaluation metrics often fall short in providing a detailed ability assessment. This problem is further exacerbated by their inability to match the given answer accurately, leading to significant robustness issues. To address these challenges, research communities have introduced a series of benchmarks including MME \cite{fu2023mme}, MMBench \cite{liu2023mmbench}, MM-Vet \cite{yu2023mmvet}, SEED-Bench \cite{li2023seedbench}, and Q-Bench \cite{wu2023qbench}. These benchmarks systematically structure and evaluate complex multi-modal tasks. We applied our model, in a zero-shot manner, to five recently popular multi-modal benchmarks. For a fair comparison, we select models with similar language model sizes, particularly those from the LLaMA family, and detail their differences in the vision encoder. The results of our evaluation are listed in Table \ref{table:zeroshot-multimodal-bench}. In the table, \modelname achieves higher zero-shot performance in terms of MMBench, MM-Vet, and Q-Bench. Conversely, the performance on MME is lower because of the limited number of test samples in MME, which could potentially lead to sensitive fluctuations in performance. Particularly, it exhibits significant improvement on Q-Bench, a benchmark for examining the low-level visual perception of MLLMs. This improvement occurs when applying a smaller visual backbone (i.e., ViT-L), leading to enhanced low-level visual perception. This demonstrates the effectiveness of our training strategy for training visual backbone.

\paragraph{Natural Language Understanding and Generation.}
\vspace{-2ex}
\begin{table}[h]
\centering
    \tablestyle{7pt}{1.1} 
    \def \w{15pt}
    \resizebox{\linewidth}{!}{
    \begin{tabular}{l|c|c|c|c|c}
        \shline
        Method & MMLU & BBH & AGIEval & ARC-c & ARC-e  \\
        \hline
        LLaMA-2 \cite{Touvron2023Llama2} & 46.8 & 38.2 & 21.8 & 40.3 & 56.1 \\
        WizardLM \cite{Xu2023WizardLM} & 38.1 & 34.7 & 23.2 & 47.5 & 59.6 \\
        LLaMA-2-Chat \cite{Touvron2023Llama2} & 46.2 & 35.6 & 28.5 & 54.9 & 71.6 \\
        Vicuna-v1.5 \cite{zheng2023vicuna} & 51.1 & 41.2 & 21.2 & 56.6 & 72.8 \\
        \hline
        \textbf{\modelname} & \textbf{53.4} & \textbf{45.0} & \textbf{32.7} & \textbf{65.8} & \textbf{79.9} \\
        \shline
    \end{tabular}
    }
    \caption{\textbf{Performance on pure-text benchmarks of \modelname} compared to LLaMA-2 (7B) family variants. We adopt 5-shot for MMLU and 0-shot for BBH, AGIEval, and ARC as \cite{2023opencompass}.}
    \label{table:text-results}
\end{table}
\vspace{-2ex}
Current MLLMs often outperform in various multi-modal downstream tasks by leveraging the power of large language models. Nevertheless, the intrinsic capabilities of these models often play a significant role in determining the performance of MLLMs, an aspect that has often been overlooked in prior multi-modal language model studies. Accordingly, we have also assessed the performance of our model in the context of natural language understanding and generation. We perform the evaluation on MMLU \cite{hendrycks2020mmlu}, BBH \cite{suzgun2022bbh}, AGIEval \cite{zhong2023agieval} and ARC \cite{clark2018arc}. The results are illustrated in Table \ref{table:text-results}. As observed in the table, \modelname excels in examination and reasoning, showing a significant improvement on MMLU and BBH by 2.3\% and 3.8\% respectively. This indicates that \modelname not only performs well on multi-modal tasks but also achieves better performance compared to the other instruction-tuned LLMs, showing the promising way for developing strong MLLMs.

\paragraph{Zero-Shot Video Question Answering.}

\begin{table}[h]
\centering
    \tablestyle{4pt}{1.1} 
    \def \w{15pt}
    \resizebox{\linewidth}{!}{
    \begin{tabular}{l|cc|cc|cc}
        \shline
        \multirow{2}{*}{Method} & \multicolumn{2}{c}{MSRVTT-QA} & \multicolumn{2}{c}{MSVD-QA} & \multicolumn{2}{c}{TGIF-QA}  \\
        \cmidrule(lr){2-3} \cmidrule(lr){4-5} \cmidrule(lr){6-7}
        ~ & Accuracy & Score & Accuracy & Score & Accuracy & Score \\
        \hline
        \multicolumn{7}{l}{\textit{\textbf{Exacting Match}}} \\
        \hline
        Flamingo-80B \cite{alayrac2022flamingo} & 17.4 & - & 35.6 & - & - & -  \\
        FrozenBiLM \cite{yang2022frozenbilm} & 16.8 & - & 32.2 & - & 41.0 & - \\
        BLIP-2 \cite{Li2023BLIP2} & 9.2 & - & 18.3 & - & - & - \\
        HiTeA \cite{ye2023hitea} & 21.7 & - & 37.4 & - & - & - \\
        InstructBLIP \cite{Dai2023InstructBLIP} & 22.1 & - & 41.8 & - & - & - \\
        \hline
        \textbf{\modelname} & \textbf{23.6} & - & \textbf{42.4} & - & \textbf{61.6} & - \\
        \hline \hline
        \multicolumn{7}{l}{\textit{\textbf{GPT-Assisted}}} \\
        \hline
        Video Chat \cite{li2023videochat} & 45.0 & 2.5 & 56.3 & 2.8 & 34.4 & 2.3 \\
        LLaMA-Adapter \cite{Gao2023LLaMAAdapterV2} & 43.8 & 2.7 & 54.9 & 3.1 & - & - \\
        Video-LLaMA \cite{Zhang2023VideoLLaMA} & 29.6 & 1.8 & 51.6 & 2.5 & - & - \\
        Video-ChatGPT \cite{Maaz2023VideoChatGPT} & \textbf{49.3} & 2.8 & 64.9 & 3.3 & 51.4 & 3.0  \\
        \hline
        \textbf{\modelname} & 46.7 & \textbf{2.9} & \textbf{65.4} & \textbf{3.5} & \textbf{67.1} & \textbf{3.7} \\
        \shline
    \end{tabular}
    }
    \caption{\textbf{Zero-shot evaluation on video question answering.} Accuracy and relevance score are reported.}
    \vspace{-2ex}
    \label{table:video-results}
\end{table}
Given that videos can be viewed as a sequence of images, we conducted a comprehensive quantitative evaluation using several commonly employed video question-answering datasets, including MSRVTT-QA \cite{xu2017msrvttqa}, MSVD-QA \cite{xu2017msrvttqa}, and TGIF-QA \cite{jang2017tgif}. These datasets aided in the zero-shot evaluation of the model's ability to understand video content, with the results summarized in Table \ref{table:video-results}. We employed two types of evaluations: 1) Exact matching, which is commonly used in previous video question-answering evaluations; and 2) GPT-assisted evaluation \cite{Maaz2023VideoChatGPT} that assesses the model's capabilities by measuring the accuracy of the model’s generated predictions and providing a relative score on a scale of 1-5. We observe that our model achieves superior results on all three video datasets under a zero-shot setting. Furthermore, in terms of relevancy, our model generates more accurate answers than other video MLLMs, thereby demonstrating its superiority and excellent generalization capabilities.

\subsection{Discussion}
\begin{figure}[t]
    \centering
    \includegraphics[width=1\linewidth]{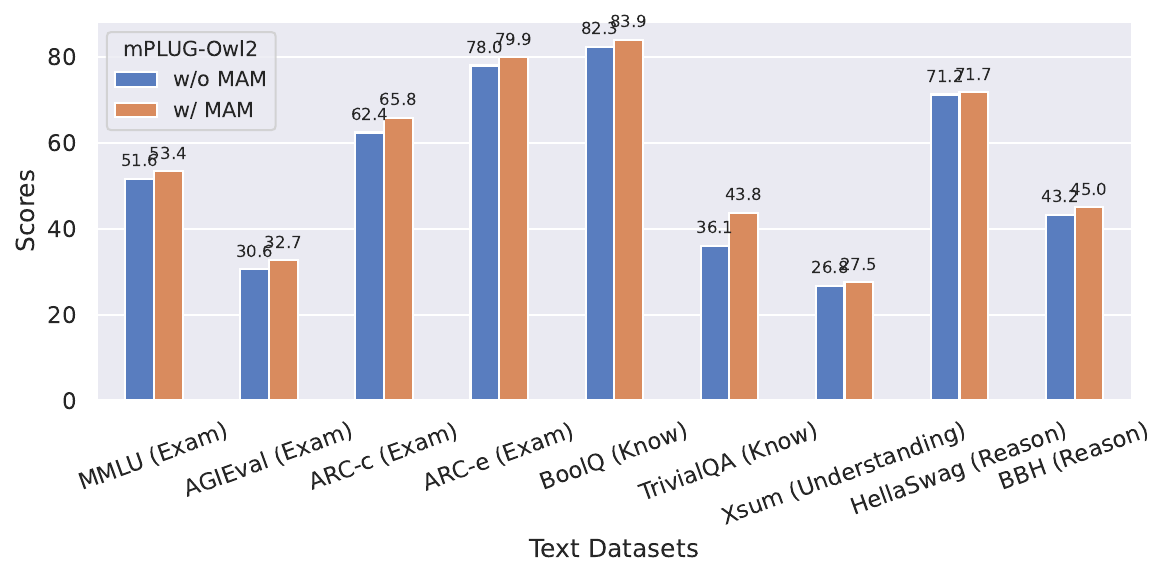}
    \vspace{-4ex}
    \caption{Performance of text benchmarks across various capabilities under modality collaboration.}
    \label{fig:text-collab}
    \vspace{-2ex}
\end{figure}

\paragraph{Modality Collaboration for Text Performance.}
To demonstrate how modality collaboration enhances not only the multi-modal performance but also the text capability of MLLMs, we evaluate the performance of text benchmarks in terms of various abilities including examination, knowledge, understanding, and reasoning. As observed in Figure \ref{fig:text-collab}, both examination and knowledge capabilities of MLLMs have significantly improved thanks to the benefits of modality collaboration facilitated by the modality-adaptive module. This improvement arises because multi-modal data allows the model to utilize visual information to understand concepts that cannot be described through language. Similarly, the model can generate richer and more substantial responses due to a more concrete understanding of these concepts. Additionally, multi-modal data enhances the reasoning ability of the model because images contain rich information (such as relationships and spatial aspects). The model learns from these aspects and associates them with the text, thereby indirectly enhancing the reasoning ability of the text.

\paragraph{Impact of Joint Vision-Language Instruction Tuning.}

\begin{table}
\centering
    \tablestyle{7pt}{1.1} 
    \def \w{15pt}
    \resizebox{\linewidth}{!}{
    \begin{tabular}{c|c|c|cccc}
        MAM  & Text Inst.   &  MM Inst.  & VQAv2  & Q-Bench & MMLU & BBH \\
        \shline
         & \checkmark &  & 58.2 & 54.4 & 51.8 & 43.6 \\
         &  & \checkmark & 76.3 & 61.3 & 45.4 & 25.7 \\
         &  \checkmark  & \checkmark  & 76.2 & 60.3 & 51.6 & 43.2  \\ \hline \hline
        
        \checkmark & \checkmark &  & 60.5 & 55.6 & 51.8 & 44.0  \\
        \checkmark &  & \checkmark & 76.5 & 60.2 & 46.1 & 30.6 \\
        \checkmark &  \checkmark  & \checkmark  & \textbf{76.8} & \textbf{62.2} & \textbf{52.8} & \textbf{45.0}  \\
    \end{tabular}
    }
    \caption{Performance comparison among different types of instruction data and structures.}
    \vspace{-2ex}
    \label{table:data-ablation}
\end{table}
Table \ref{table:data-ablation} presents the results of instruction tuning with various types of data as well as whether using modality-adaptive module. These results show that even without multi-modal instruction data, the model's performance on multi-modal benchmarks is respectable due to the effective vision-language alignment achieved during pre-training. However, when solely using multi-modal instruction data, we observe an increase in performance on multi-modal datasets, while performance on text tasks decreases by about 5.7\%. This phenomenon can be counterbalanced by the joint vision-language tuning proposed, as shown in the table's third row, where the multi-modal performance begins to slightly decrease due to modality interference. To counter this drawback, we apply our proposed modality-adaptive module to the model. Results show that the performance on both multi-modal and text benchmarks improves, with a minimum increase of 0.6\% on the VQAv2 dataset and 1.6\% on MMLU.

\paragraph{Impact of Trainable Vision Encoder.}
\begin{table}
\centering
    \tablestyle{7pt}{1.1} 
    \def \w{15pt}
    \resizebox{1\linewidth}{!}{
    \begin{tabular}{c|c|cccc}
        Unfreeze  &  Layer-wise lr. & VQAv2 & TextVQA & MMBench & Q-Bench \\
        \shline
         &  & 74.8 & 39.8 & 63.8 & 60.7  \\
        \checkmark &  & 76.2 \textbf{\color{red}{(+1.4)}} & 40.3 \textbf{\color{red}{(+0.5)}} & 62.7 \textbf{\color{deepgreen}{(-1.1)}} & 61.6 \textbf{\color{red}{(+0.9)}}  \\
        \checkmark  & \checkmark  & 76.8 \textbf{\color{red}{(+2.0)}}  & 42.5 \textbf{\color{red}{(+2.7)}} & 64.5 \textbf{\color{red}{(+0.7)}} & 62.2 \textbf{\color{red}{(+1.5)}}   \\
    \end{tabular}
    }
    \caption{Influence of learning strategies for visual encoder.}
    \vspace{-2ex}
    \label{table:vit-ablation}
\end{table}
Table \ref{table:vit-ablation} delivers the performance of the training vision encoder during instruction tuning with modality collaboration. It can be observed that enabling the vision encoder to be trainable improves performance on VQAv2 and Q-Bench by at least 1.4\% and 0.9\%, respectively, suggesting the benefits of modality collaboration. Conversely, it results in a 1.1\% performance drop in MM-Bench, indicating a degree of forgetting and damage to the general visual representation due to the limited diversity of instruction data. To mitigate this challenge, we apply layer-wise learning rate decay with an exponential decay factor of 0.9, which preserves the representation of lower layers while modifying higher semantic representations. By applying the layer-wise learning rate decay, we can notice that performance on TextVQA has increased further with 2.2\%, showing the effectiveness of our training strategy.

\paragraph{Impact of Number of Learnable Queries.}
\begin{table}
\centering
    \tablestyle{8pt}{1.1}
    \def \w{8pt}
    \resizebox{\linewidth}{!}{
    \begin{tabular}{c|cccc}
        \# Learnable Queries  &  VQAv2 & TextVQA & MMBench & Q-Bench \\
        \shline
        8  & 58.3 & 18.6 & 47.6 & 52.4  \\
        16  & 66.2 & 28.5 & 52.9 & 54.9  \\
        32  & 72.4 & 36.3 & 60.2 & 57.8 \\
        64  & \textbf{76.8} & 42.5 & \textbf{64.5} & \textbf{62.2}  \\
        128  & 76.7 & \textbf{44.4} & 63.6 & 61.6  \\
    \end{tabular}
    }
    \caption{Performance in terms of number of learnable queries.}
    \vspace{-2ex}
    \label{table:query-ablation}
\end{table}
To investigate the effect of the number of learnable queries $\mathcal{Q}$, we conduct experiments using different numbers of queries in the visual abstractor, as shown in Table \ref{table:query-ablation}. It can be observed that the model consistently exhibits improvement as the number of learnable queries increases until it reaches a saturation point, suggesting that 64 may be the optimal number for representing an image. Notably, there is a significant performance boost observed when the number is increased from 8 to 64, e.g., the performance of VQAv2 is increased 18.5\%. These findings suggest that a higher number of learnable queries can capture image information more comprehensively, thereby enhancing the model's image comprehension capabilities.

\begin{table}[htpb]
\centering
    \tablestyle{7pt}{1.1} 
    \def \w{8pt}
    \resizebox{\linewidth}{!}{
    \begin{tabular}{c|ccccc}
        Resolution &  VQAv2 & TextVQA & MMBench & MM-Vet & Q-Bench \\
        \shline
        $224 \times 224$  & 76.8 & 42.5 & 64.5 & 34.0 & 62.2  \\
        $336 \times 336$  & 78.5 \textbf{\color{red}{(+1.7)}} & 49.8 \textbf{\color{red}{(+7.3)}} & 65.2 \textbf{\color{red}{(+0.7)}} & 34.6 \textbf{\color{red}{(+0.6)}} & 62.4 \textbf{\color{red}{(+0.2)}} \\
        $448 \times 448$  & 79.4 \textbf{\color{red}{(+2.6)}} & 54.3 \textbf{\color{red}{(+11.8)}}  & 65.4 \textbf{\color{red}{(+0.9)}} & 36.2 \textbf{\color{red}{(+2.2)}} & 62.6 \textbf{\color{red}{(+0.4)}}  \\
    \end{tabular}
    }
    \caption{Influence of different input image resolutions.}
    \vspace{-2ex}
    \label{table:resolution-ablation}
\end{table}
\paragraph{Impact of Image Resolution.}
Image resolution plays a crucial role in vision-language tasks, as a higher resolution can reduce image blur and improve understanding of fine-grained details. To explore the impact of image resolution on performance across different benchmarks, we adjust the image resolution from $224\times 224$ to $448 \times 448$ and the results are listed in Table \ref{table:resolution-ablation}. 
As observed in the table, using a higher resolution proves advantageous for multi-modal tasks, particularly in the question answering scenario. Specifically, the performance of VQAv2 has increased from 76.8 to 79.4, representing a 2.6\% boost. Simultaneously, there is an 11.8 point lift in the TextVQA benchmark when enlarging the resolution from $224\times 224$ to $448\times 448$. This suggests that OCR-related tasks benefit significantly from increasing the resolution.

\subsection{Qualitative Analysis}
\begin{figure}[h]
    \centering
    \includegraphics[width=\linewidth]{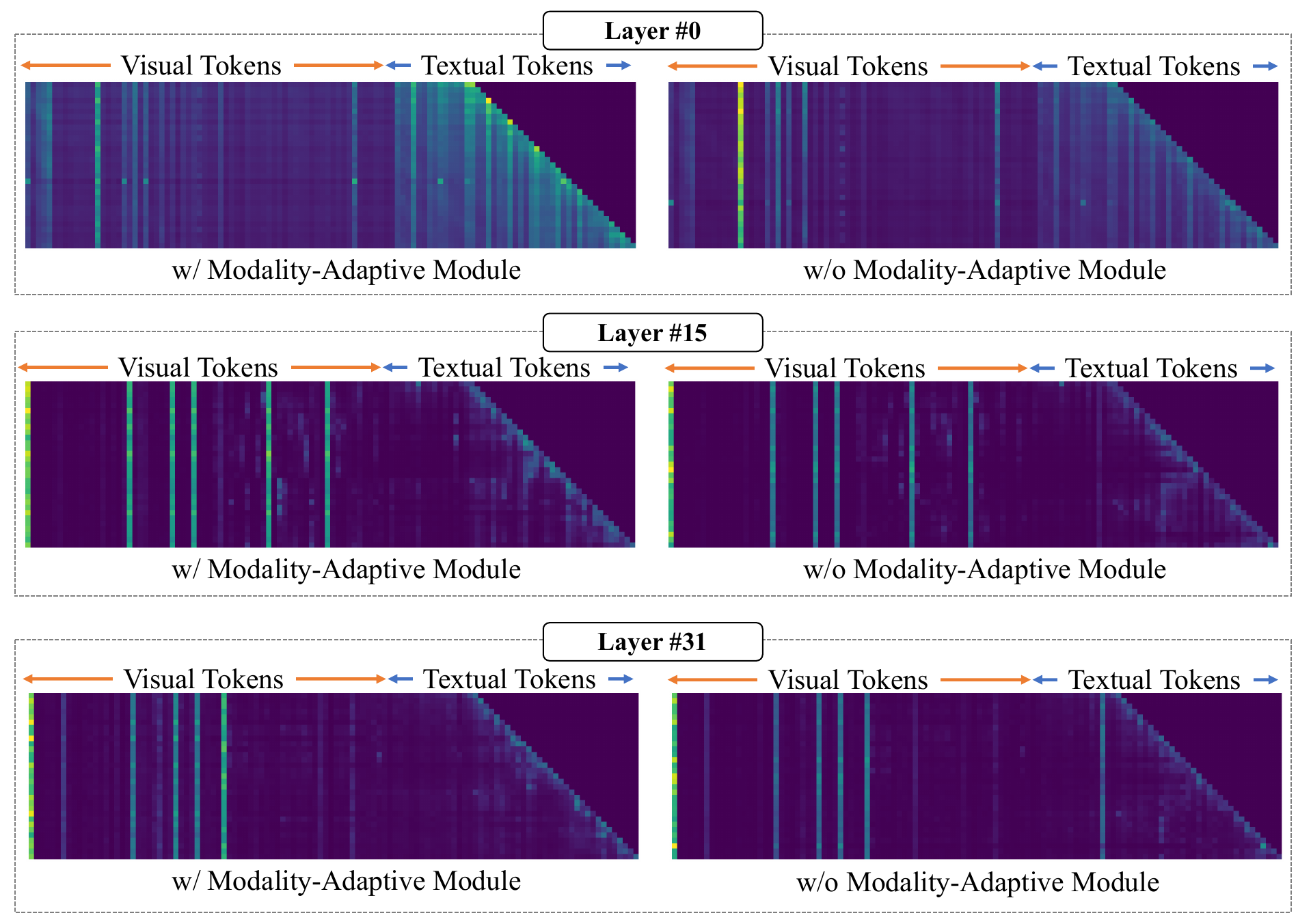}
    \caption{Visualization of the attention maps with and without the Modality-Adaptive Module. We demonstrate the attention maps for the 0-th, 15-th, and 31-st layers, where the range of visual tokens is indicated by orange and the range of text tokens is indicated by blue.}
    \vspace{-2ex}
    \label{fig:visumap}
\end{figure}
\paragraph{Impact of Modality-Adaptive Module in Multi-Modal Scenario.}


We investigate the impact of the Modality-Adaptive Module in multi-modal scenarios by visualizing the attention maps of \modelname with and without this module using image caption input, as shown in \Cref{fig:visumap}. Each attention map illustrates the attention scores of generated tokens on the input sequence during the generation process.

It can be observed that regardless of whether the Modality-Adaptive Module is incorporated or not, the model focuses more on the textual tokens in the earlier layers while paying more attention to the visual tokens in the later layers. This suggests that the modeling of visual and textual information plays different roles in the collaboration of multi-modal language models (MLLMs). An intuitive explanation is that MLLMs initially use syntactic information to comprehend instructions and then identify relevant visual content tokens by considering the textual input.

When using the Modality-Adaptive Module, it can be observed that the model explicitly pays more attention to the textual content in the earlier stages and focuses more on the visual content in the later stages. The Modality-Adaptive Module prevents visual and textual tokens from being treated as the same and encourages collaboration between different modalities.

\paragraph{Impact of Modality-Adaptive Module in Unrelated-Modality Scenarios.}
\begin{figure}[h]
    \centering
    \includegraphics[width=\linewidth]{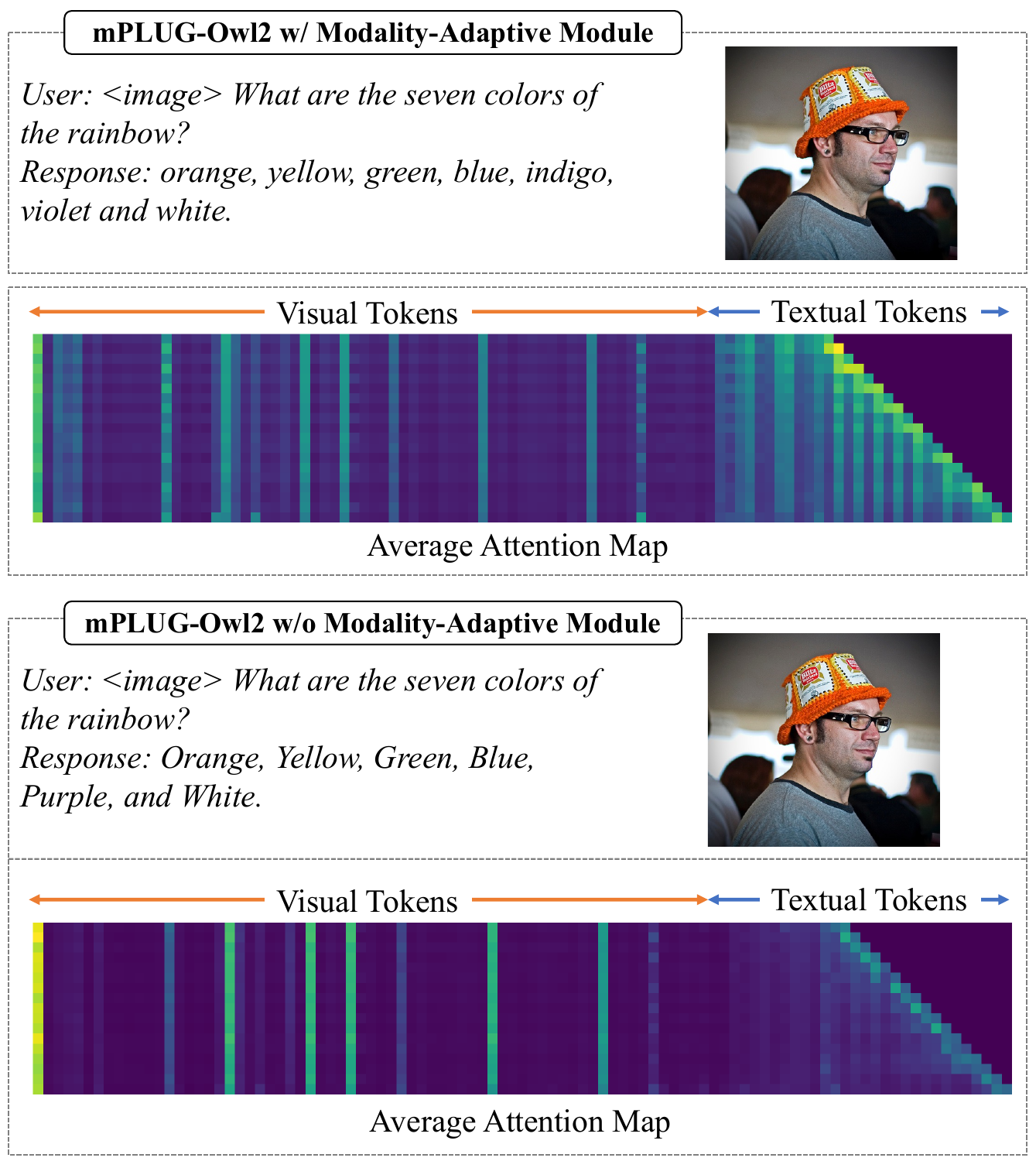}
    \caption{Visualization of the attention maps with and without the Modality-Adaptive Module. We demonstrate the average of attention maps across each layer, where the range of visual tokens is indicated by orange and the range of text tokens is indicated by blue.}
    \vspace{-2ex}
    \label{fig:modal-unrelated}
\end{figure}
We present a question: "What are the seven colors of the rainbow?" along with a randomly selected image. In this example, the image input acts as a disturbance to the model. We aim to investigate the impact of our module on data that contains unrelated modalities. The responses and attention maps of the model are shown in \Cref{fig:modal-unrelated}. Our proposed model, \modelname, which incorporates the Modality-Adaptive Module, accurately identifies all seven colors. During the generation process, it can be observed that the model primarily focuses on the textual input. On the other hand, when the Modality-Adaptive Module is not utilized, \modelname only identifies six colors. The model's ability to comprehend text instructions is disrupted, and it is also evident that it places more emphasis on the image during generation. Thanks to the Modality-Adaptive Module, \modelname is better able to capture modality-specific features when modeling multimodal inputs. This enhances the adaptability of modality collaboration, resulting in reduced disturbance when the text and image are unrelated.

\section{Conclusion}
In this paper, we present \modelname, a highly capable generalist model by leveraging modality collaboration for enhancing performance across both text and multi-modal tasks. The inclusion of shared functional modules and a modality-adaptive module in \modelname strengthens the model's ability to harmonize modality collaboration and preserve modality-specific characteristics. The extensive experimental evaluations highlight \modelname's proficiency in generalizing across various tasks, thereby achieving state-of-the-art performances with a singular, generalized model. Most notably, \modelname stands as the first MLLM model to exhibit the phenomena of modality collaboration in both pure-text and multi-modal contexts. This not only enhances the model's vision-language understanding but also improves its language capabilities in terms of understanding, knowledge, and reasoning. This represents a significant contribution to the field and opens up exciting opportunities for the future development of multi-modal foundation models.

{
    \small
    \bibliographystyle{ieeenat_fullname}
    \bibliography{main}
}

\clearpage
\appendix

\section{Additional Experimental Results}
In this section, we provide more experimental results for the completeness of our proposed method.


\subsection{Hallucination Evaluation}
\begin{figure}[h]
    \centering
    \includegraphics[width=\linewidth]{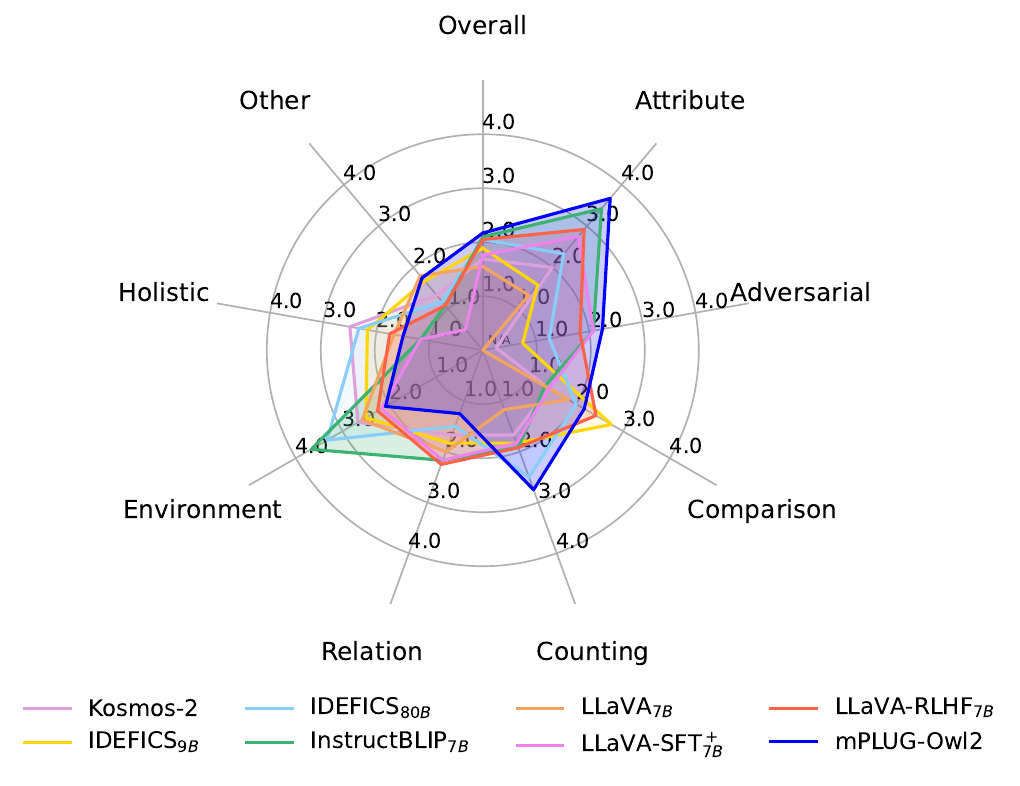}
    \caption{Detailed performance of various models across the eight categories in MMHal-Bench \cite{Sun2023LLavaRlhf}, where "Overall" represents the average performance across all categories.}
    \label{fig:mmhal-bench}
    \vspace{-2ex}
\end{figure}

We measure the hallucination of our model on image description using MMHal-Bench \cite{Sun2023LLavaRlhf} and compare the results with other recent vision-language models, including Kosmos-2 \cite{Peng2023Kosmos2GM}, IDEFICS \cite{laurencon2023idefics}, InstructBLIP \cite{Dai2023InstructBLIP}, LLaVA \cite{Liu2023Llava}, and LLaVA-RLHF \cite{Sun2023LLavaRlhf}. Following \cite{Sun2023LLavaRlhf}, we use GPT-4 to evaluate the overall score and hallucination rate of different MLLMs. As depicted in Figure \ref{fig:mmhal-bench}, we find that our \modelname tends to generate the response with reduced hallucination compared to other methods, especially surpassing IDEFICS \cite{laurencon2023idefics} with 80 billion parameters, showing the superiority of our methods. Besides, we can notice that our model excels at attribute and counting because the visual abstractor can effectively identify the main parts of the image, which reduces the hallucination.

\begin{figure*}[htp]
    \centering
    \includegraphics[width=0.6\linewidth]{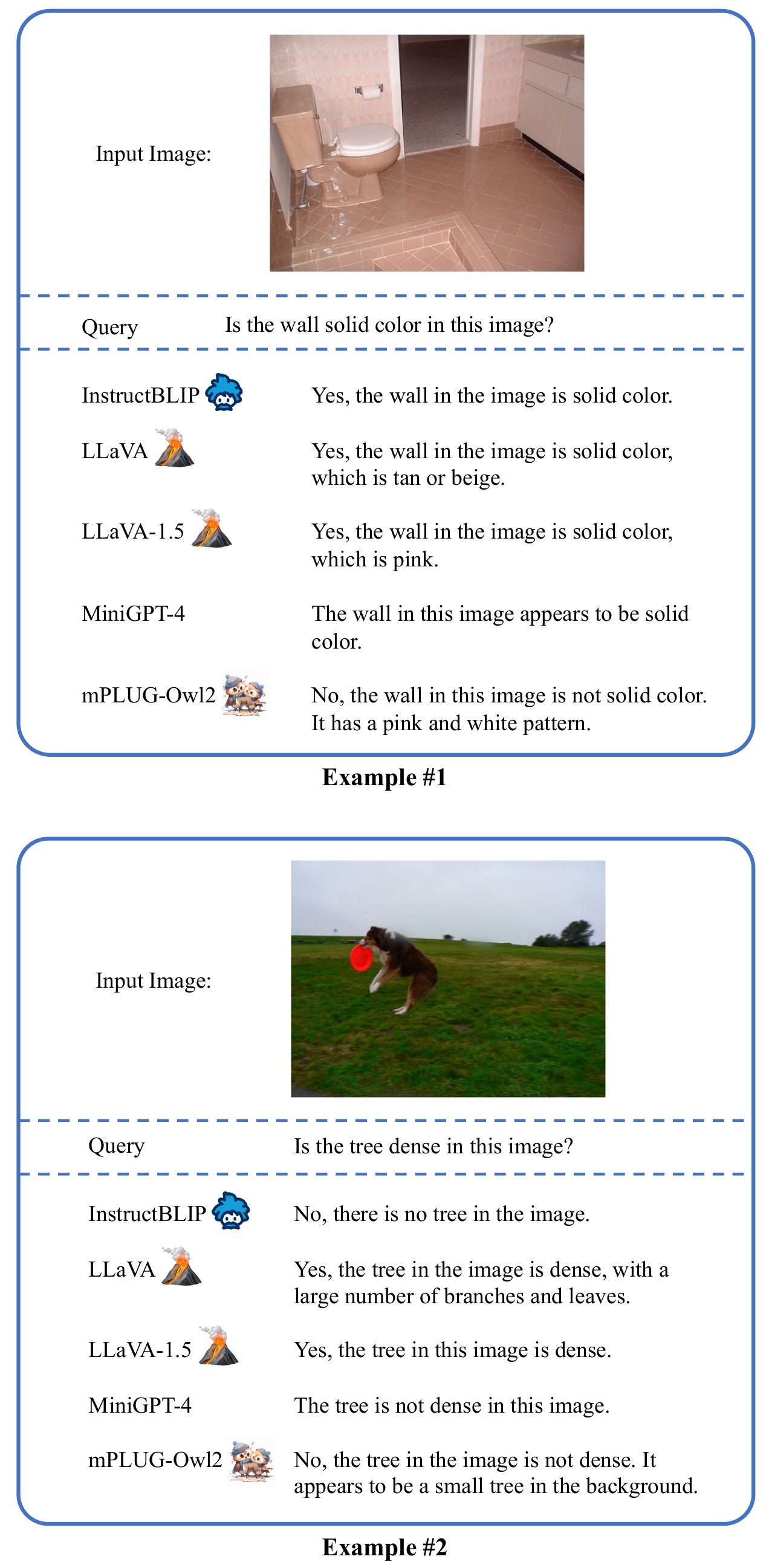}
    \caption{Example cases compared with InstructBLIP~\cite{Dai2023InstructBLIP}, LLAVA~\cite{Liu2023Llava}, LLAVA-1.5~\cite{Liu2023LLava15}, MiniGPT-4~\cite{Zhu2023MiniGPT4} and our~\modelname.}
    \label{fig:case}
\end{figure*}

We also study the hallucination of recent popular MLLMs and present the results in \Cref{fig:case}. In the first example, the query asks the models to recognize the pattern on the wall. However, the pattern is not clearly visible in the image, causing other models to mistakenly perceive it as a solid color. Our model, on the other hand, accurately notices the white pattern on the wall and correctly answers the question. In the second example, there are only a few trees in the image. However, InstructBLIP incorrectly considers that there are no trees in the image. LLaVA and LLaVA-1.5, on the other hand, hallucinate and consider the tree in the image to be dense. MiniGPT-4 gives the correct answer, but with minimal explanation. Our \modelname, however, answers the question correctly and provides a more detailed explanation.

\subsection{POPE Evaluation}
We also conduct the hallucination evaluation using POPE \cite{Li2023pope}, the results are shown in Table \ref{tab:pope_results}. As we can observe in the table, we can find \modelname achieves higher F1 scores on the popular and adversarial split, showing the robustness of our model in terms of object hallucination compared to other MLLMs.
\begin{table*}[h!]
\centering
\caption{\textbf{Object hallucination benchmark using POPE evaluation pipeline }. "Yes" signifies the likelihood of the model producing a positive response.
}
\label{tab:pope_results}
\resizebox{\textwidth}{!}{%
\begin{tabular}{l|l|ccccccc}
\toprule
Datasets & Metrics & \modelname & Shikra \cite{Chen2023Shikra} & InstructBLIP \cite{Dai2023InstructBLIP}  & MiniGPT-4 \cite{Zhu2023MiniGPT4} & LLaVA \cite{Liu2023Llava} & MM-GPT \cite{gong2023mmgpt} & mPLUG-Owl \cite{ye2023mplugowl} \\
\cmidrule(lr){1-2}\cmidrule(lr){3-9}
\multirow{5}{*}{Random}
& Accuracy ($\uparrow$)       & 88.28 & 86.90 & 88.57 & 79.67 &50.37  & 50.10& 53.97 \\
& Precision ($\uparrow$)      & 94.34 & 94.40 & 84.09 &78.24  &50.19  & 50.05&52.07\\
& Recall ($\uparrow$)         & 82.20 & 79.27 & 95.13 &82.20  &  99.13& 100.00&99.60  \\
& F1-Score ($\uparrow$)       & 87.85 & 86.19 & \textbf{89.27}  &80.17  & 66.64 &  66.71 &68.39 \\
& Yes ($\rightarrow 50\%$)    & 44.91 & 43.26  & 56.57 & 52.53 & 98.77 &  99.90&95.63 \\
\cmidrule(lr){1-2}\cmidrule(lr){3-9}
\multirow{5}{*}{Popular}
& Accuracy ($\uparrow$)       & 86.20 & 83.97  & 82.77 &69.73  &49.87  & 50.00&50.90  \\
& Precision ($\uparrow$)      & 89.46 & 87.55 & 76.27 & 65.86 &49.93  & 50.00&50.46  \\
& Recall ($\uparrow$)         & 82.06 & 79.20  & 95.13 &81.93  & 99.27 & 100.00&99.40  \\
& F1-Score ($\uparrow$)       & \textbf{85.60} & 83.16 & 84.66 & 73.02 & 66.44 & 66.67 & 66.94\\
& Yes ($\rightarrow 50\%$)   & 45.86 & 45.23  & 62.37 & 62.20 & 99.40 &100.00  &98.57\\
\cmidrule(lr){1-2}\cmidrule(lr){3-9}
\multirow{5}{*}{Adversarial}
& Accuracy ($\uparrow$)       & 84.12 & 83.10  & 72.10  &65.17  &  49.70& 50.00 & 50.67\\
& Precision ($\uparrow$)      & 85.54 & 85.60  & 65.13 & 61.19 & 49.85 & 50.00 & 50.34\\
& Recall ($\uparrow$)         & 82.13 & 79.60 & 95.13 & 82.93 & 99.07 & 100.00 & 99.33\\
& F1-Score ($\uparrow$)       & \textbf{83.80} & 82.49 & 77.32 &  70.42& 66.32 &66.67  & 66.82\\
& Yes ($\rightarrow 50\%$)    & 48.00 & 46.50 & 73.03 &67.77  & 99.37 &   100.00&98.67\\
\bottomrule
\end{tabular}%
}
\end{table*}

\subsection{Detailed Evaluation Results on MMBench}
MMBench \cite{liu2023mmbench} is a meticulously designed benchmark that comprehensively assesses the diverse skills of vision-language models. The results from the test set for various MLLMs are presented in Table \ref{tab:mmbench_results}.

\begin{table*}[!t]
\small
\centering
\begin{tabular}{l|cc|cccccccc}
\toprule
Method & Language Model & Vision Model & Overall & LR & AR & RR & FP-S & FP-C & CP \\
\midrule
MMGPT \cite{gong2023mmgpt} & LLaMA-7B & CLIP ViT-L/14 & 16.0 & 1.1 & 23.8 & 20.7 & 18.3 & 5.2 & 18.3 \\
MiniGPT-4 \cite{Zhu2023MiniGPT4} & Vicuna-7B & EVA-G & 12.0 & 13.6 & 32.9 & 8.9 & 28.8 & 11.2 & 28.3 \\
InstructBLIP \cite{Dai2023InstructBLIP} & Vicuna-7B & EVA-G & 33.9 & 21.6 & 47.4 & 22.5 & 33.0 & 24.4 & 41.1 \\
LLaMA-Adapter-v2 \cite{Gao2023LLaMAAdapterV2} & LLaMA-7B & CLIP ViT-L/14 & 38.9 & 7.4 & 45.3 & 19.2 & 45.0 & 32.0 & 54.0 \\
LLaVA \cite{Sun2023LLavaRlhf} & Vicuna-7B & CLIP ViT-L/14 & 36.2 & 15.9 & 53.6 & 28.6 & 41.8 & 20.0 & 40.4 \\
G2PT \cite{liu2023mmbench} & Vicuna-7B & ViT-G & 39.8 & 14.8 & 46.7 & 31.5 & 41.8 & 34.4 & 49.8 \\
Otter-I \cite{Li2023Otter} & LLaMA-7B & CLIP ViT-L/14 & 48.3 & 22.2 & 63.3 & 39.4 & 46.8 & 36.4 & 60.6 \\
mPLUG-Owl$^\dag$ \cite{ye2023mplugowl} & LLaMA-7B & CLIP ViT-L/14 & 62.3 & \textbf{37.5} & \textbf{75.4} & 56.8 & \textbf{67.3} & 52.4 & 67.2 \\
Shikra \cite{Chen2023Shikra} & Vicuna-7B & CLIP ViT-L/14 & 60.2 & 33.5 & 69.6 & 53.1 & 61.8 & 50.4 & 71.7 \\
\midrule 
\modelname & LLaMA2-7B & CLIP ViT-L/14 & \textbf{65.4} & 29.2 & 69.7 & \textbf{61.7} & 67.0 & \textbf{60.0} & \textbf{79.5} \\
\bottomrule
\end{tabular}%
\caption{CircularEval multi-choice accuracy results on MMBench \cite{liu2023mmbench} dev set. We adopt the following
abbreviations: LR for Logical Reasoning; AR for Attribute Reasoning; RR for Relation Reasoning; FP-C for Fine-grained Perception (Cross Instance); FP-S for Fine-grained Perception (Single Instance); CP for Coarse Perception. Baseline results are taken from \cite{liu2023mmbench}. $^\dag$ denotes the model is carefully optimized for MMBench.}
\label{tab:mmbench_results}
\end{table*}

\subsection{Detailed Evaluation Results on MM-Vet}
We provide the detailed results of MM-Vet in Table \ref{tab:mmvet_benchmark}. It can be observed that by training the visual encoder of \modelname, it exhibits stronger OCR capability compared to the model with the same backbone (i.e., LLaVA, Otter). Besides, \modelname surpasses models with stronger language decoders such as LLaVA-13B which equips LLM with 13 billion parameters.

\begin{table*}[t]
  \centering
    \begin{tabular}{lcccccccc}
    \shline
        Model & Rec & OCR & Know & Gen & Spat & Math & Total  \\ 
        \hline
        Transformers Agent (GPT-4) \cite{OpenAI2023gpt4} & 18.2 & 3.9 & 2.2 & 3.2 & 12.4 & 4.0 & 13.4$\pm$0.5 \\
        MiniGPT-4-7B \cite{Zhu2023MiniGPT4} & 27.4 & 15.0 & 12.8 & 13.9 & 20.3 & 7.7 & 22.1$\pm$0.1  \\ 
        BLIP-2-12B \cite{Li2023BLIP2} &  27.5 & 11.1 & 11.8 & 7.0 & 16.2 & 5.8 & 22.4$\pm$0.2 \\ 
        LLaVA-7B \cite{Liu2023Llava} & 28.0 & 17.1 & 16.3 & 18.9 & 21.2 &  11.5 & 23.8$\pm$0.6 \\ 
        MiniGPT-4-13B \cite{Zhu2023MiniGPT4} & 29.9 & 16.1 & 20.4 & 22.1 & 22.2 & 3.8 & 24.4$\pm$0.4 \\
        Otter-9B \cite{Li2023Otter} & 27.3 & 17.8 & 14.2 & 13.8 & 24.4 & 3.8 & 24.7$\pm$0.3 \\ 
        OpenFlamingo-9B \cite{awadalla2023openflamingo} & 28.7 & 16.7 & 16.4 & 13.1 & 21.0 & 7.7 & 24.8$\pm$0.2 \\
        InstructBLIP-13B \cite{Dai2023InstructBLIP} & 30.8 & 16.0 & 9.8 & 9.0 & 21.1 & 10.5 & 25.6$\pm$0.3 \\
        InstructBLIP-7B \cite{Dai2023InstructBLIP} & 32.4 & 14.6 & 16.5 & 18.2 & 18.6 & 7.7 & 26.2$\pm$0.2 \\
        LLaVA-7B (LLaMA-2) \cite{Liu2023Llava} & 32.9 & 20.1 & 19.0 & 20.1 & 25.7 & 5.2 & 28.1$\pm$0.4 \\
        LLaMA-Adapter v2-7B \cite{Gao2023LLaMAAdapterV2} &  38.5 & 20.3 &  \textbf{31.4} &  \textbf{33.4} & 22.9 & 3.8 & 31.4$\pm$0.1 \\
        LLaVA-13B (V1.3) \cite{Liu2023Llava} &  38.1 & 22.3 & 25.2 & 25.8 &  \textbf{31.3} & 11.2 &  32.5$\pm$0.1 \\
        LLaVA-13B (LLaMA-2) \cite{Liu2023Llava} &  39.2 &  22.7 &  26.5 &  29.3 & 29.6 & 7.7 &  32.9$\pm$0.1 \\
        \hline
        \modelname &  \textbf{41.3} &  \textbf{27.4} &  27.5 & 27.9 & 30.3 & 7.7 &  \textbf{36.2}$\pm$0.1 \\
        \shline
    \end{tabular}
    \caption{Evaluation results on various MLLMs regarding each core VL capability on MM-Vet \cite{yu2023mmvet}. Rec stands for recognition; Know indicates knowledge; Gen is generation; Spat means spatial. All the numbers are presented in \% and the full score is 100\%.}
    \label{tab:mmvet_benchmark}
\end{table*}

\subsection{Detailed Evaluation Results on Q-Bench}
For evaluating the low-level visual perception abilities, we have included the results of Q-Bench \cite{wu2023qbench} on the test set. By training the visual encoder, the ability of \modelname in terms of low-level perception has been improved significantly, as it outperforms the model with a stronger visual encoder (i.e., ViT-L (0.3B) v.s. ViT-G (1.9B)), showing the effectiveness of our training paradigm.

\begin{table*}[h]
\centering
\resizebox{\textwidth}{!}{%
\begin{tabular}{l|cccccccc}
\shline
\textbf{Method}  & \textbf{Yes-or-no} & \textbf{What} & \textbf{How} & \textbf{Distortion} & \textbf{Others} & \textbf{In-context Distortion} & \textbf{In-context Others} & \textbf{Overall} \\
\hline
IDEFICS \cite{laurencon2023idefics}         & 0.6004             & 0.4642        & 0.4671       & 0.4038              & 0.5990          & 0.4726                         & 0.6477                     & 0.5151           \\
InstructBLIP \cite{Dai2023InstructBLIP}     & 0.7099             & 0.5141        & 0.4300       & 0.4500              & 0.6301          & 0.5719                         & 0.6439                     & 0.5585           \\
Kosmos-2 \cite{Peng2023Kosmos2GM}         & 0.6058             & 0.3124        & 0.3539       & 0.3865              & 0.4654          & 0.4349                         & 0.4735                     & 0.4334           \\
LLaMA-Adapter-v2 \cite{Gao2023LLaMAAdapterV2} & 0.6661             & 0.5466        & 0.5165       & \textbf{0.5615}              & 0.6181          & 0.5925                         & 0.5455                     & 0.5806           \\
LLaVA-1.5 \cite{Liu2023LLava15}        & 0.6460             & \textbf{0.5922}        & 0.5576       & 0.4798              & 0.6730          & 0.5890                         & \textbf{0.7376}                     & 0.6007           \\
LLaVA \cite{Liu2023Llava}           & 0.5712             & 0.5488        & 0.5185       & 0.4558              & 0.5800          & 0.5719                         & 0.6477                     & 0.5472           \\
MiniGPT-4 \cite{Zhu2023MiniGPT4}       & 0.6077             & 0.5033        & 0.4300       & 0.4558              & 0.5251          & 0.5342                         & 0.6098                     & 0.5177           \\
mPLUG-Owl \cite{ye2023mplugowl}        & 0.7245             & 0.5488        & 0.4753       & 0.4962              & 0.6301          & \textbf{0.6267}                         & 0.6667                     & 0.5893           \\
Otter \cite{Li2023Otter}          & 0.5766             & 0.3970        & 0.4259       & 0.4212              & 0.4893          & 0.4760                         & 0.5417                     & 0.4722           \\
Qwen-VL \cite{Bai2023QwenVL}         & 0.6533             & 0.6074        & 0.5844       & 0.5413              & 0.6635          & 0.5822                         & 0.7300                     & 0.6167           \\
Shikra  \cite{Chen2023Shikra}          & 0.6909             & 0.4793        & 0.4671       & 0.4731              & 0.6086          & 0.5308                         & 0.6477                     & 0.5532          \\
\hline
\modelname           & \textbf{0.7318}             & 0.5531        & \textbf{0.5864}       & 0.5374              & \textbf{0.7136}          & 0.5788                         & 0.7338                     & \textbf{0.6294}          \\
\shline
\end{tabular}
}
\caption{Detailed evaluation results for different MLLMs on the test set of Q-Bench \cite{wu2023qbench}.}
\label{tab:qbench-eval}
\end{table*}

\subsection{Detailed Evaluation Results on MMHal-Bench}
We include Table~\ref{tab:mmhalbench_eval} for the full evaluation results on MMHal-Bench \cite{Sun2023LLavaRlhf}.
\begin{table*}[h]
\centering
\resizebox{\textwidth}{!}{%
\begin{tabular}{l|cc|cccccccc}
\shline
\multirow{2}{*}{\textbf{Method}} & \textbf{Overall} & \textbf{Hallucination} & \multicolumn{8}{c}{\textbf{Score in Each Question Type} $\uparrow$} \\
& \textbf{Score} $\uparrow$ & \textbf{Rate} $\downarrow$ & {Attribute} & {Adversarial} & {Comparison} & {Counting} & {Relation} & {Environment} & {Holistic} & {Other} \\
\hline
Kosmos-2 \cite{Peng2023Kosmos2GM} & 1.69 & 0.68 & 2.00 & 0.25 & 1.42 & 1.67 & 1.67 & 2.67 & \textbf{2.50} & 1.33 \\
IDEFICS$_\textsc{9B}$ \cite{laurencon2023idefics} & 1.89 & 0.64 & 1.58 & 0.75 & 2.75 & 1.83 & 1.83 & 2.50 & 2.17 & 1.67 \\
IDEFICS$_\textsc{80B}$ \cite{laurencon2023idefics} & 2.05 & 0.61 & 2.33 & 1.25 & 2.00 & 2.50 & 1.50 & 3.33 & 2.33 & 1.17 \\
InstructBLIP$_\textsc{7B}$ \cite{Dai2023InstructBLIP} & 2.10 & 0.58 & 3.42 & 2.08 & 1.33 & 1.92 & 2.17 & 3.67 & 1.17 & 1.08 \\
InstructBLIP$_\textsc{13B}$ \cite{Dai2023InstructBLIP} & 2.14 & 0.58 & 2.75 & 1.75 & 1.25 & 2.08 & \textbf{2.50} & \textbf{4.08} & 1.50 & 1.17 \\
LLaVA$_\textsc{7B}$ \cite{Liu2023Llava} & 1.55 & 0.76 & 1.33 & 0.00 & 1.83 & 1.17 & 2.00 & 2.58 & 1.67 &  \textbf{1.83} \\
LLaVA-RLHF$_\textsc{7B}$ \cite{Sun2023LLavaRlhf} & 2.05 & 0.68 & 2.92 & 1.83 & \textbf{2.42} & 1.92 & 2.25 & 2.25 & 1.75 & 1.08 \\
\hline
\modelname & \textbf{2.17} & \textbf{0.56} & \textbf{3.67} & \textbf{2.25} & 2.17 & \textbf{2.75} & 1.25 & 2.08 & 1.50 & 1.75 \\
\shline
\end{tabular}%
}
\caption{Detailed evaluation results for different MLMMs on MMHal-Bench.}
\label{tab:mmhalbench_eval}
\end{table*}

\section{Implementation}
\subsection{Data Mixture}
In this section, we detail our final training data mixture used during the instruction tuning stage in Table \ref{tab:data_mixture}. Specifically, we process the VQAv2 \cite{balanced_vqa_v2} data by selecting the answer with the highest confidence and combining question-answer pairs that share the same image. This combining strategy is also applied to GQA \cite{hudson2019gqa}, OKVQA \cite{marino2019okvqa}, and OCRVQA \cite{mishra2019ocrvqa} datasets. Additionally, for multiple-choice questions in A-OKVQA \cite{schwenk2022aokvqa}, we augment the dataset by switching the order of options to enhance robustness in terms of multiple choices. For caption datasets like COCO \cite{lin2014coco} and TextCaps \cite{sidorov2020textcaps}, we randomly select one caption from the ground truth for each image. Concurrently, some regional-VQA \cite{yu2016refcoco, krishna2017visualgenome} datasets are also used to improve regional abilities.
\begin{table}[!htbp]
\small
    \centering
    \begin{tabular}{l|ll}
         \shline
         \textbf{Data Type} & \textbf{Data Name} & \textbf{Size} \\
         \hline
         \multirow{2}{*}{Text} & ShareGPT \cite{ShareGPT2023} & 40K \\
         & SlimOrca \cite{SlimOrca} & 518K \\ \hline
         Dialogue & LLaVA \cite{Liu2023Llava} & 158K \\ \hline
         \multirow{2}{*}{Caption} & COCO \cite{lin2014coco} & 82K \\
         & TextCaps \cite{sidorov2020textcaps} & 22K \\ \hline
         \multirow{5}{*}{VQA} & VQAv2 \cite{balanced_vqa_v2} & 83K \\
         & GQA \cite{hudson2019gqa} & 72K \\
         & OKVQA \cite{marino2019okvqa} & 9K \\
         & OCRVQA \cite{mishra2019ocrvqa} & 80K \\
         & A-OKVQA \cite{schwenk2022aokvqa} & 50K \\ \hline
         \multirow{2}{*}{Regional-VQA} & RefCOCO \cite{yu2016refcoco} & 30K \\
         & VisualGenome \cite{krishna2017visualgenome} & 86K \\ \hline
         \multicolumn{2}{c}{Total} & 1.23M  \\
         \shline 
    \end{tabular}
    \caption{Instruction-following Data Mixture of \modelname.}
    \label{tab:data_mixture}
\end{table}

\subsection{Training Hyper-parameters}
\begin{table}[htbp]
    \centering
    \tablestyle{7pt}{1.0}
    \begin{tabular}{l|cc}
         \shline \\
         Configuration            & Pre-training & Instruction Tuning \\
         \hline
         ViT init.                & CLIP-L/14 \cite{radford2021clip} & Pre-train stage \\
         LLM init.                & LLaMA-2 \cite{Touvron2023Llama2} & LLaMA-2 \cite{Touvron2023Llama2} \\
         Visual Abstractor init.  & Random & Pre-train stage \\
         Image resolution         & $224\times 224$ & $448\times 448$ \\
         ViT sequence length      & 256 & 1024 \\
         LLM sequence length      & 256 & 2048\\
         Learnable query numbers  & 64 & 64\\
         Optimizer                & \multicolumn{2}{c}{AdamW} \\
         Optimizer hyperparameter & \multicolumn{2}{c}{$\beta_{1}=0.9, \beta_{2}=0.98, \epsilon=1e^{-6}$} \\
         Peak learning rate       & $1e^{-4}$ & $2e^{-5}$ \\
         Minimum learning rate    & $1e^{-6}$ & $1e^{-7}$ \\
         ViT learning rate decay  & \multicolumn{2}{c}{0} \\
         ViT Drop path rate       & \multicolumn{2}{c}{0} \\
         Learning rate schedule   & \multicolumn{2}{c}{Cosine} \\
         Weight decay             & 0.05 & 0 \\
         Gradient clip            & \multicolumn{2}{c}{1.0} \\
         Training steps           & 42,500 & 4,800 \\
         Warm-up steps            & 1,000 & 250 \\
         Global batch size        & 8,192 & 256 \\
         Gradient Acc.            & \multicolumn{2}{c}{16} \\
         Numerical precision      & \multicolumn{2}{c}{$\mathtt{bfloat16}$} \\
         Optimizer sharding       & \multicolumn{2}{c}{\checkmark} \\
         Activation checkpointing & \multicolumn{2}{c}{\checkmark} \\
         Model parallelism        & 1 & 2 \\
         Pipeline parallelism     & \multicolumn{2}{c}{1} \\
         \shline
    \end{tabular}
    \caption{Training hyper-parameters of \modelname.}
    \label{tab:hyperparam}
\end{table}

We report the detailed training hyper-parameter settings of \modelname in Table~\ref{tab:hyperparam}. Specifically, we leverage the model parallelism with Megatron \cite{shoeybi2019megatron} distributed training framework to ensure a larger resolution training while maintaining efficiency.

\section{Summary of the Evaluation Benchmarks}
\label{app:benchmark}

We provide a detailed summary of the used evaluation benchmarks and corresponding metrics in Table~\ref{tab:benchmark}.

\begin{table*}[ht]
    \centering
    \resizebox{\textwidth}{!}{
    \begin{tabular}{l|l|l|l|l}
        \toprule
         Task & Dataset & Description & Split & Metric  \\
         \midrule
         \multirow{2}{*}{Image Caption} & COCO & Captioning of natural images & karpathy-test & CIDEr ($\uparrow$) \\
         & Flickr30K & Captioning of natural images & karpathy-test & CIDEr ($\uparrow$) \\
         \midrule
         \multirow{6}{*}{General VQA} & VQAv2 & VQA on natural images & test-dev & VQA Score ($\uparrow$) \\
         & OKVQA & VQA on natural images requiring outside knowledge & val & VQA Score ($\uparrow$) \\
         & GQA & VQA on scene understanding and reasoning & test-balanced & EM ($\uparrow$) \\
         & VizWizQA & VQA on photos taken by people who are blind & test-dev & VQA Score ($\uparrow$)\\
         & TextVQA & VQA on natural images containing text & val & VQA Score ($\uparrow$) \\
         & SciQA-Img & Multi-choice VQA on a diverse set of science topics & test & Accuracy ($\uparrow$) \\
         \midrule
         \multirow{3}{*}{VideoQA} & MSRVTT-QA & Video Question Answering & test &  Accuracy  ($\uparrow$) / Relevance Score ($\uparrow$) \\
         & MSVD-QA & Video Question Answering & test &  Accuracy  ($\uparrow$) / Relevance Score ($\uparrow$) \\
         & TGIF-QA & GIF Question Answering & test &  Accuracy  ($\uparrow$) / Relevance Score ($\uparrow$) \\
         \midrule
         \multirow{5}{*}{Text Benchmark} & MMLU & A benchmark designed to measure knowledge acquirement & dev &  Accuracy ($\uparrow$) \\
         & BBH & A suite of 23 challenging BIG-Bench tasks & test & Accuracy ($\uparrow$) \\
         & AGIEval & A human-centric benchmark specifically designed to evaluate the general abilities of foundation model & test &  Accuracy ($\uparrow$) \\
         & ARC-c & A multiple-choice question-answering dataset, containing questions from science exams from grade 3 to grade 9. & test &  Accuracy ($\uparrow$) \\
         & ARC-e & A multiple-choice question-answering dataset, containing questions from science exams from grade 3 to grade 9. & test &  Accuracy ($\uparrow$) \\
         \midrule
         \multirow{5}{*}{Instruction Following} & MME & Open-ended VL Benchmark by yes/no questions & Perception & Accuracy ($\uparrow$) \\
          & MMBench & Open-ended VL Benchmark by Multi-choice VQA with Circular Evaluation & test & Accuracy ($\uparrow$) \\
          & MM-Vet & Open-ended VL Benchmark with Various Abilities & test & GPT-$4$ Score ($\uparrow$) \\
          & SEED-Bench & Open-ended VL Benchmark by Multi-choice VQA & Image \& Video & Accuracy ($\uparrow$) \\
          & Q-Bench & Open-ended Low-level Vision Benchmark by Multi-choice VQA & test & Accuracy ($\uparrow$) \\
         \midrule
         \multirow{2}{*}{Hallucination} & POPE & Object existence by yes/no questions & random/popular/adversarial & Accuracy / Precision / Recall / F1 ($\uparrow$) \\
         & MMHal-Bench & Open-ended hallucination benchmarks & test & GPT-$4$ Score ($\uparrow$) \\
         \bottomrule
    \end{tabular}
    }
    \caption{Summary of the evaluation benchmarks of \modelname. EM stands for exacting matching.}
    \label{tab:benchmark}
\end{table*}

\section{Broader Impact}
\modelname employs off-the-shelf LLM and web-sourced data. Consequently, it inherits some of the weaknesses of the original LLM and web-crawled data, such as generating uncensored text or producing biased outputs. We address these shortcomings by enhancing the model's grounding on the visual and instructional input and executing joint vision-language instruction tuning on a diverse range of high-quality datasets. However, we advise against deploying \modelname models for any downstream applications without prior evaluation of safety and fairness specific to the respective application.

\end{document}